\pdfoutput=1

\documentclass[11pt]{article}

\usepackage[final]{acl}

\usepackage{times}
\usepackage{latexsym}

\usepackage[T1]{fontenc}

\usepackage[utf8]{inputenc}

\usepackage{microtype}

\usepackage{inconsolata}

\usepackage{graphicx}
\usepackage{booktabs}
\usepackage{tabularx}
\usepackage{subcaption}

\title{Does Synthetic Data Help 
\\ Named Entity Recognition for Low-Resource Languages?}

\author{
    Gaurav Kamath\thanks{Work done during an internship at the National Research Council, Canada.}\textsuperscript{1, 2} \quad
    Sowmya Vajjala\textsuperscript{3}
    \\[1ex]
    \textsuperscript{1}McGill University, Canada \quad
    \textsuperscript{2}Mila - Quebec AI Institute, Canada \\
    \textsuperscript{3}National Research Council, Canada
    \\[1ex]
        \texttt{gaurav.kamath@mail.mcgill.ca} \quad
        \texttt{sowmya.vajjala@nrc-cnrc.gc.ca}
    }

\begin{document}
\maketitle
\begin{abstract}
 We explore whether synthetic datasets generated by large language models using a few high quality seed samples are useful for low-resource named entity recognition, considering 11 languages from three language families. Our results suggest that synthetic data created with such seed data is a reasonable choice when there is no available labeled data, and is better than using entirely automatically labeled data. However, a small amount of high-quality data, coupled with cross-lingual transfer from a related language, always offers better performance.\footnote{Data and code available at: \url{https://github.com/grvkamath/low-resource-syn-ner}.}
\end{abstract} 

\section{Introduction}
\label{sec:intro}
Named Entity Recognition (NER) for low-resource languages aims to produce robust systems for languages with limited labeled training data available, and has been an area of increasing interest within natural language processing (NLP) over the past decade. Two common approaches to address this data scarcity are cross-lingual transfer and data augmentation/synthesis; recent research has in particular explored the usefulness of large language models (LLMs) for such data augmentation and synthetic data creation in NLP \cite{whitehouse-etal-2023-llm,li2023synthetic}, while their use for NER is also emerging \cite{bogdanov2024nuner,dao-etal-2025-overcoming}. 

\begin{figure}[htb!]
  \centering
  \includegraphics[width=0.85\linewidth]{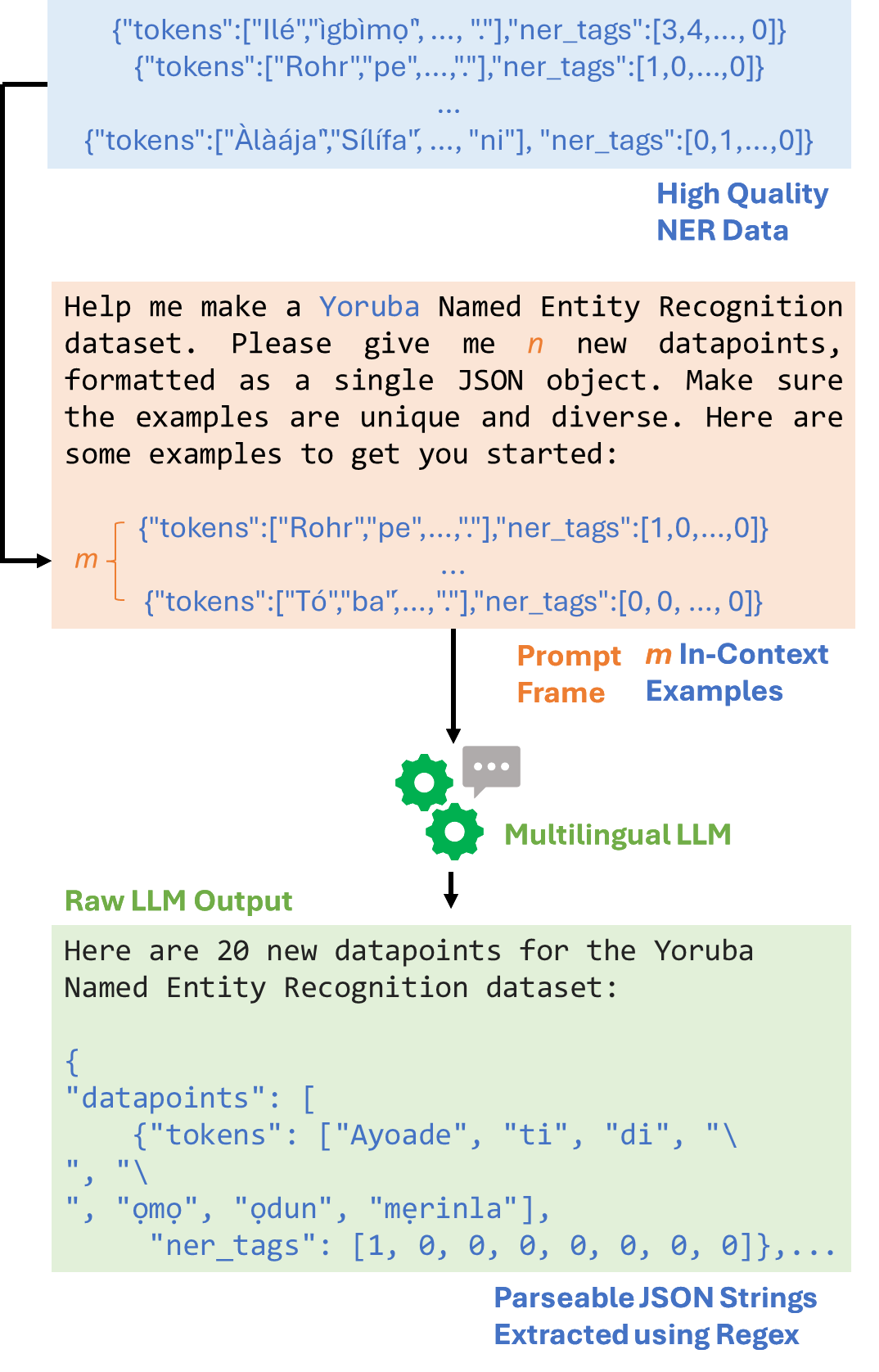}
  \caption {High-level overview of our data generation process. We use multilingual large language models to generate new NER data points on the basis of a handful of high quality human labeled data points. See Section \ref{subsec:datagen} for more.}
  \label{fig:datagen}
\end{figure}

In this background, we propose LLM-based synthetic data generation using a small amount of gold examples (Figure~\ref{fig:datagen}) as an alternative to relying on automatically created datasets for low-resource NER. With experiments covering 11 languages from 3 language families\textemdash Danish, Swedish and Slovak from the Indo-European language family; Swahili, Kinyarwanda, Yoruba and Igbo from the Niger-Congo language family, and Kannada, Malayalam, Tamil, and Telugu from the Dravidian language family, we show that:
\begin{enumerate}\itemsep0em
\item Even a small amount of human annotated data can yield far better performance than much larger amounts of synthetic data.
\item Zero-shot transfer from a related language can provide high baselines for low-resource language NER.
\item Synthetic data generated by prompting an LLM with a few high quality (generally human labeled) examples (Figure~\ref{fig:datagen}) could be better than using automatically labeled datasets when training low-resource NER models. 
\end{enumerate} 

We start with a review of related literature (Section~\ref{sec:related}) and describe our data generation approach and experimental setup in Section~\ref{sec:methods}, followed by a discussion of the results (Section~\ref{sec:results}), limitations (Section~\ref{sec:limitations}) and broader impact (Section~\ref{sec:ethics}). 

\section{Related Work}
\label{sec:related}

NER in low resource settings has long been a topic of interest in NLP. Significant research examines cross-lingual transfer from a high resource source language to a lower-resource target language for the task \cite{rahimi-etal-2019-massively,mueller-etal-2020-sources,zeng-etal-2022-dualner,zhao-etal-2022-transadv,yang-etal-2022-crop,zhou-etal-2022-conner}, while other approaches have explored the creation of synthetic datasets through e.g. parallel corpora or machine translation \cite{mayhew-etal-2017-cheap, ni-etal-2017-weakly,pan-etal-2017-cross,xie-etal-2018-neural,liu-etal-2021-mulda,yang-etal-2022-crop,fetahu-etal-2022-dynamic}. There are also large existing automatically constructed multilingual NER datasets that rely on sources such as Wikipedia \cite{pan-etal-2017-cross,krishnan2021employing,malmasi2022multiconer}, some of which have become a part of large multilingual benchmarks \cite{asai-etal-2024-buffet}. 
 
More recent work has explored using LLMs as data generators for NER \cite{bogdanov2024nuner,heng-etal-2024-proggen,evuru-etal-2024-coda}. We build on such work, but differ from their methods. Our data generation process uses high quality, human validated examples as seeds, and we not only evaluate different LLMs (both open and closed-source) as synthetic data generators, but also experiment with 11 languages covering three language families and five base scripts. To our knowledge, this is the first attempt to explore using large language models for synthetic data generation in low-resource NER, and the first to cover $>10$ languages.

\section{Our Approach}
\label{sec:methods}  
At a high level, our approach involves two steps:
\begin{enumerate}
\item Using the train split of a high quality (usually manually annotated) NER dataset for a target language to generate synthetic data for that language with the help of an LLM (Section~\ref{subsec:datagen}); and then 
\item Comparing the performance of an NER model on the test split of the high quality dataset when trained on synthetic data from Step 1 and another model trained on the train split of the same high quality dataset (Section~\ref{subsec:training}).
\end{enumerate}

\subsection{Synthetic Data Generation: }
\label{subsec:datagen}
Our synthetic data generation process (shown in Figure \ref{fig:datagen}) involves using LLMs to generate new synthetic data points on the basis of existing, high quality NER annotations as described below: 
\begin{itemize}
\item First, we randomly sample $m$ data points from the train split of an organic (i.e. non-synthetic) NER dataset. 
\item Next, we format and append these data points to a prompt asking the model to produce $n$ new, unique data points on the basis of the $m$ data points in the prompt. 
\item We submit this prompt as input to the LLM, and extract the correctly-formatted data points from its response; 
\item We repeat steps (1)-(3) $k$ times, with each call to the model choosing a different random sample of organic data points.  
\end{itemize}
In our experiments, we set $m$ to 10, $n$ to 20, and $k$ to 500. This sets an upper cap of 5000 synthetic training data points, if every model response contains perfectly formatted data points. We present and solicit data structured as JSON strings to the LLMs, and extract well-formatted samples from model responses using regular expressions. Appendix \ref{app:datagen} provides further details about this process.  

\begin{table*}[]
    \centering
\resizebox{\textwidth}{!}{
    \begin{tabular}{@{}lr@{}}
    \toprule
       LLM Response Quality  &  Examples \\ \midrule
        Well-Formatted  & \texttt{\{"data": [\{
      "tokens": ["Lars", "Løkke", "Rasmussen", "besøgte", "firmaet", "i", "Odense", "."]}, \\
       & \texttt{"ner\_tags": [1, 2, 2, 0, 0, 0, 5, 0]\}}, ... \\ \midrule
        Unequal Token \& Tag Lengths  & \texttt{\{"id": "4123", "tokens": ["Wananchi", "wamekunja", "mashitaka", "."],} \\ 
          & \texttt{"ner\_tags": [0, 0, 0, 0, 0, 0]\}} \\ \midrule
        Run-On \& Incomplete Data  &  \texttt{\{"id": "9000", "tokens": ["Olorun", "l\d{\`{e}}\d{\`{e}}", ..., "\`{o}", "\`{o}", "\`{o}", "\`{o}", "\`{o}", "\`{o}", "\`{o}", "\`{o}", "\`{o}", "\`{o}", "\`{o}", "\`{o}", "\`{o}"} \\
         & \texttt{\{"id":"4617","tokens":["\d{O}\d{\'{d}}\`{n}", "\d{\`{o}}s\`{e}", "l\`{e}", "\d{\`{o}}r\d{\'{i}}", "-", "\`{e}d\`{e}", "\d{O}\d{\'{b}}\d{\'{f}}emi", "\`{a}f\'{u}", "f\`{u}n", "\`{a}w\d{o}n", "gb", "." ]} \\ 
         & \texttt{"ner\_tags":[8,0,0,0,0} \\ \midrule
        Empty Responses \& Prompt Continuations  & \texttt{<EOS\_TOKEN>} \\
         & \texttt{<EOS\_TOKEN>include a mix of names, locations, organizations...} \\
        \bottomrule
    \end{tabular}
    }
    \caption{Examples of different types of responses from the synthetic data-generating LLMs tested, across languages}
    \label{tab:responsequality}
\end{table*}

We compare three LLMs as our source of synthetic data: \texttt{GPT-4.1}\footnote{We use \texttt{gpt-4.1-2025-04-14}. Note that in an earlier draft of this work, we used \texttt{gpt-4-turbo} \cite{achiam2023gpt}, when it represented the state-of-the-art; surprisingly, \texttt{gpt-4-turbo} yielded slightly better results. Nevertheless, here we report results on \texttt{GPT-4.1}, to better represent currently available models.} \cite{openai2025gpt4_1}, which we assume to be the state of the art; \texttt{Llama-3.1-8B-Instruct} \cite{dubey2024llama}, as a much smaller, open-source instruction-tuned model; and finally, \texttt{aya-expanse-32b} \cite{dang2024aya}, as a larger open source multilingual LLM.   

\subsection{Training NER models: }
\label{subsec:training}
For all experiments, we use the pre-trained version of \texttt{XLM-RoBERTa-large} \cite{conneau-etal-2020-unsupervised} as our base model and fine-tune it on our synthetic and organic training sets in two distinct settings.
\begin{enumerate}
\item In the first setting, we use our synthetic data to train an NER model from scratch, by fine-tuning \texttt{XLM-RoBERTa-large} on target language NER data.
\item In the second setting, we first fine-tune the model on the high quality NER data in a \textit{related} source language\footnote{See Table \ref{tab:relatedlangappendix} in Appendix \ref{app:appendix_relatedlanguagemodel} for the full list of chosen related languages for all the target languages.}, and then further fine-tune this NER model using our synthetic or organic target language data.
\end{enumerate}

While the first setting\textemdash which we name \textsc{ner from scratch}\textemdash aims to shed light on the relative utility of synthetic data for training an NER model (largely) from the ground up, the latter \textemdash which we name \textsc{ner fine-tuning}\textemdash simulates a common setting, when a lower resource language lacks adequate NER data, but is related to a higher-resource language with existing NER systems. In both settings, we modulate the amount of data (both synthetic and organic) used, so as to compare model performance when trained on smaller or larger amounts of each type of data.

\paragraph{Languages \& Datasets: } We focus on 11 languages from three distinct language families: Tamil, Kannada, Malayalam, Telugu (Dravidian), Kinyarwanda, Swahili, Igbo, Yoruba (Niger-Congo), Swedish, Danish and Slovak (Indo-European). Of these, Igbo, Yoruba, and Kinyarwanda are not among the 100 languages in the \texttt{XLM-Roberta} pre-training corpus. We use the Universal NER dataset \cite{mayhew2024universal} as our high quality, manually annotated dataset for Swedish, Danish and Slovak; MasakhaNER2 \cite{adelani2022masakhaner} for Kinyarwanda, Swahili, Igbo and Yoruba; and the Naamapadam dataset \cite{mhaske-etal-2023-naamapadam} for Tamil, Kannada, Malayalam and Telugu. 

While the first two datasets are completely manually annotated, the train and validation splits of the Naamapadam dataset are constructed using parallel corpora, and thus contain some noise. Nevertheless, we choose it as our organic dataset, as (i) its test sets, which contain 500-1000 datapoints per language, are completely manually annotated, and (ii) it remains the largest NER resource for these four languages. Crucially, all of these datasets cover largely identical NER categories, allowing for comparisons between them. The Universal NER and Naamapadam datasets cover persons, locations and organizations as categories; the MasakhaNER2 data covers these three categories, as well as dates. 

Additionally, we compare models trained entirely on LLM-generated data with those trained using WikiANN \cite{pan-etal-2017-cross, rahimi-etal-2019-massively}, a large, automatically created NER dataset based on Wikipedia cross-linking, as it covers the 11 languages we study. This dataset represents a different form of synthetic data\textemdash one generated not from LLMs, but instead from scraping knowledge bases without any seed data. Although the dataset has no manual annotations, it is frequently used as a standard low-resource NER benchmark \cite{schmidt-etal-2022-slicer,asai-etal-2024-buffet}.

\section{Results}
\label{sec:results}

\begin{figure*}[htb!]
  \centering
  \begin{subfigure}{0.8\linewidth}
    \centering
    \includegraphics[width=\linewidth]{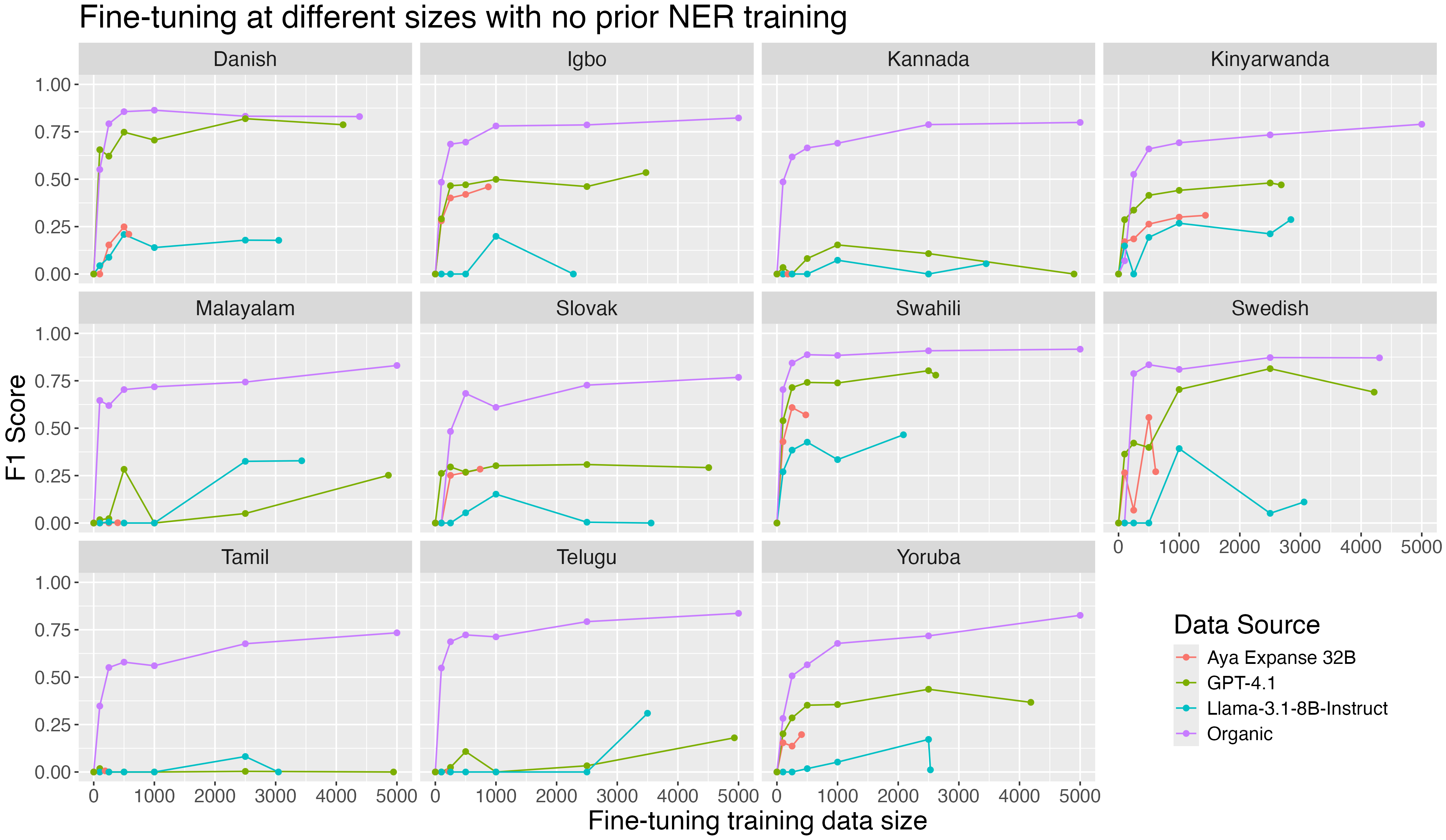}
    \caption{\textsc{ner from scratch} Setting}
    \label{fig:nfs_size}
  \end{subfigure}
  \\[1em]
  \begin{subfigure}{0.8\linewidth}
    \centering
    \includegraphics[width=\linewidth]{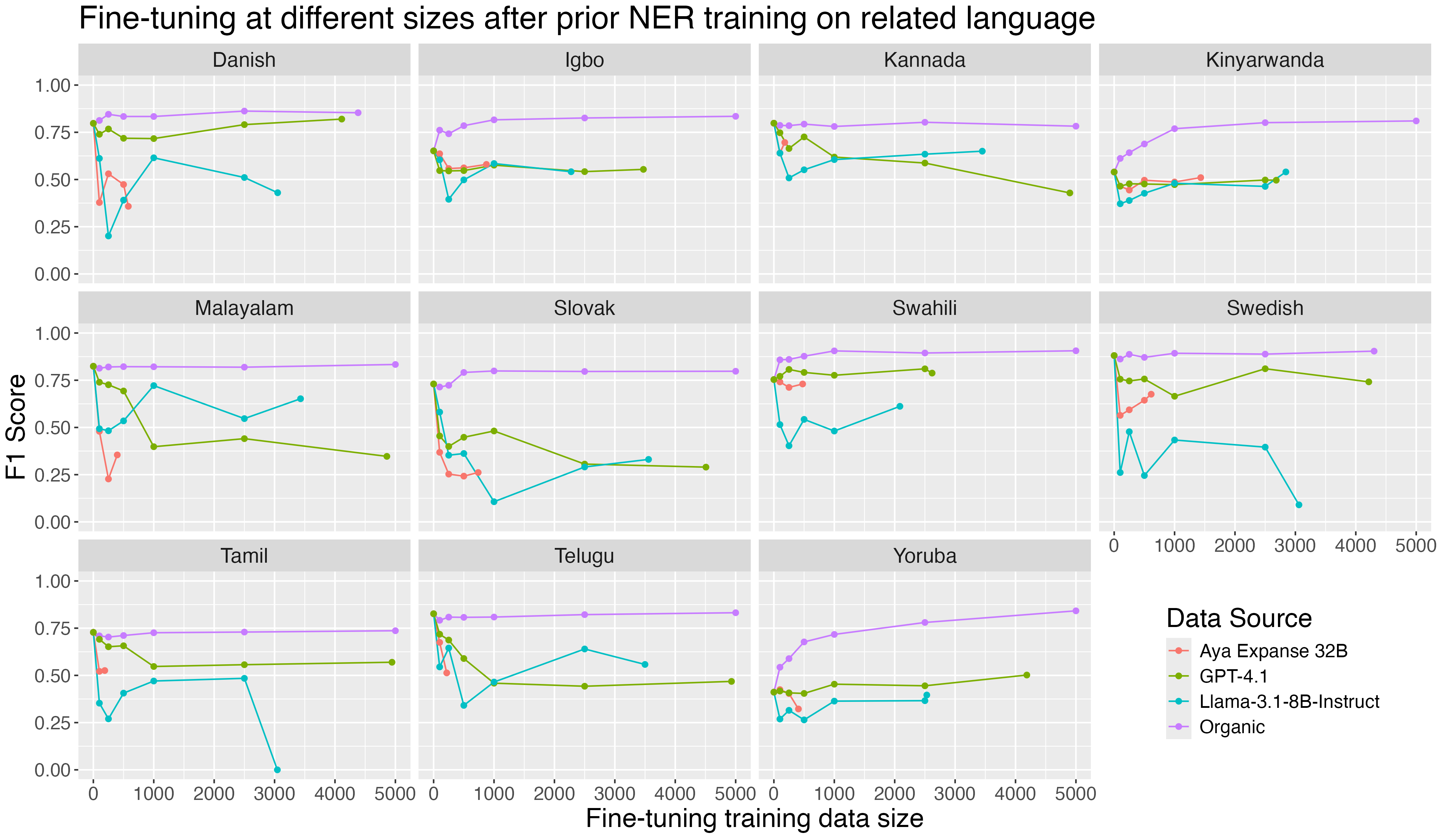}
    \caption{\textsc{ner fine-tuning} Setting}
    \label{fig:nft_size}
  \end{subfigure}
  \caption{NER model performance when trained on increasingly large subsets of training data. \texttt{aya-expanse-32b} and \texttt{Llama-3.1-8B-Instruct} produced lower amounts of usable data; this is why they do often not extend as far as organic or \texttt{GPT-4.1}-produced data in fine-tuning data size. In the \textsc{ner fine-tuning} setting, performance at \texttt{Fine-tuning Training Data Size = 0} indicates zero-shot performance of a related-language NER model.}
  \label{fig:data_size}
\end{figure*}

\subsection{Synthetic Data Generation}
\label{subsec:qualitative}

 We generate the synthetic datasets following the process described in Section~\ref{subsec:datagen}. While model responses from \texttt{GPT-4.1} are generally usable, we found more recurring errors in responses from the other two models. Some of these errors are described in Table \ref{tab:responsequality}; we discard such instances when compiling our synthetic datasets from model responses. The average percentage of usable training datapoints from \texttt{GPT-4.1}, \texttt{Llama-3.1} and \texttt{aya-expanse} are 82.6\%, 59.7\% and 11.1\% respectively. We assess the overall quality and viability of this synthetic data by measuring the performance of an NER model on a high quality, manually-annotated test set, when trained on the synthetic data.

\subsection{Training on Synthetic Data}
\label{subsec:synthdataresults}
Figure \ref{fig:data_size} shows our results when using synthetic data from different models, in both the \textsc{ner from scratch} and \textsc{ner fine-tuning} settings. While the models trained on organic data in the \textsc{ner from scratch} setting always perform better than synthetic data based models, we find that the models trained on \texttt{GPT-4.1}-generated data often come the closest to models trained on organic data compared to the other synthetic data sources. Best results with synthetic data based training are seen for Danish, followed by Swahili and Swedish. We also find that more synthetic data is not necessarily useful; for most languages, we see relative saturation after 1000 data points, and in the case of Kannada, we actually notice a drop in performance with more data. 

Perhaps more surprisingly, in the \textsc{ner fine-tuning} setting, we notice that zero-shot transfer from a related language usually outperforms the same models after they have been further fine-tuned on synthetic target language data. Fine-tuning the related language NER model with organic data from the target language helped for only Kinyarwanda and Yoruba.  This suggests that in some cases where an NER model for a related language exists, synthetic data in target languages may actually be detrimental to overall performance. Models trained on synthetic data from \texttt{GPT-4.1} do better than those trained on synthetic data from \texttt{Llama-3.1-8B-Instruct} only about half of the time; on the other hand, there are often too few usable \texttt{aya-expanse-32b} datapoints for a fair comparison. 

\paragraph{Comparison with WikiAnn: } In addition to comparing models fine-tuned on synthetic data versus organic data, we also looked into the question of whether our synthetic data generation approach offers any benefits over automatically labeled datasets, taking WikiAnn as an example.  Training models on WikiANN data leads to higher performance than training on \texttt{GPT-4.1}-generated data only for the four Dravidian languages in our data, but generally leads to significantly worse performance than training on synthetic data from \texttt{GPT-4.1} for the remaining languages (see Table~\ref{tab:appendix_wikiann} in Appendix \ref{app:wikianndetails} for detailed results). This holds in both the \textsc{ner from scratch} and \textsc{ner fine-tuning} settings, when data size is comparable; and, in the case of Danish and Swedish, training on WikiANN leads to worse performance even when there is several times more WikiANN data than \texttt{GPT-4.1}-generated data. Overall, we can conclude that using synthetic data following our approach appears to be better than relying on WikiAnn for most languages. This echoes the findings by \citet{lignos-etal-2022-toward}, who arrive at similarly negative findings around the data quality of WikiANN, and calls for not considering results on WikiANN as a benchmark for multilingual NER comparisons in the future. 


\section{Conclusions and Discussion}
\label{sec:disc}
Our results lead us to three main conclusions around the utility of LLM-generated synthetic data for low resource language NER.

\begin{enumerate}\itemsep 0ex
\item A small amount of carefully annotated data yields better performance than a large amount of synthetic data. As is evident in Figure \ref{fig:data_size}, even 100 manually annotated data points can yield NER models that cannot be matched by models trained on much larger amounts of synthetic data.

\item In many cases, zero-shot transfer from a related-language NER model is a high baseline, and that further training such a model on synthetic data may even lower the performance. We find this to be true in the case of all languages tested except the Yoruba and Swahili. For these two languages, it is worth noting that the overall baselines are lower, presumably because these languages are all lower resource than the others tested. This may explain why synthetic data yields performance gains over the zero-shot baseline, though it does not change the trend of a small amount of manually annotated data yielding far better performance.

\item Despite the fact that it falls short of manually annotated data, LLM-generated data often still yields better model performance than WikiANN, which is automatically extracted from Wikipedia texts.
\end{enumerate}

Overall, while showing how synthetic data from LLMs can help train NER models from scratch for low resource languages, our results reinforce the need for manually annotated gold test sets in benchmarking NER for lower resource languages. 

\section{Limitations}
\label{sec:limitations}
Although we experimented with many languages, the nature of the NER datasets used is relatively simple, containing only three or four entity categories (persons, locations, organizations and dates). 
Thus, we don't know if the general conclusions, especially about the quality of synthetic data, will extend to scenarios where there are many entity categories. While we did study datasets covering more than one language family, the selection of language is far from extensive, and is also constrained by the availability of human labeled test data. The observations need not necessarily hold true across all language families, naturally. Finally, to keep the experiments under control, we explored a limited set of methods for fine-tuning and synthetic data generation. Our findings should be viewed after taking these aspects into consideration.  

\section{Ethics and Broader Impact}
\label{sec:ethics}
We used publicly available datasets with human-annotated and automatically labeled data, and also created synthetically generated datasets as a part of this work. The models built using such artificially created datasets should always be validated with a human-labeled data. We did not involve any human participants in this study. All the code and generated datasets is provided at this GitHub repository to support reproducible research: \url{https://github.com/grvkamath/low-resource-syn-ner}.

\section*{Acknowledgments}
 This research was conducted at the National Research Council of Canada, thereby establishing a copyright belonging to the Crown in Right of Canada, that is, to the Government of Canada. Gaurav Kamath is supported by a Doctoral Training Award from the \textit{Fonds de Recherche du Qu\'{e}bec—Soci\'{e}t\'{e} et Culture}.

\bibliography{latex/acl_latex}

\begin{thebibliography}{32}
\providecommand{\natexlab}[1]{#1}

\bibitem[{Achiam et~al.(2023)Achiam, Adler, Agarwal, Ahmad, Akkaya, Aleman, Almeida, Altenschmidt, Altman, Anadkat et~al.}]{achiam2023gpt}
Josh Achiam, Steven Adler, Sandhini Agarwal, Lama Ahmad, Ilge Akkaya, Florencia~Leoni Aleman, Diogo Almeida, Janko Altenschmidt, Sam Altman, Shyamal Anadkat, et~al. 2023.
\newblock Gpt-4 technical report.
\newblock \emph{arXiv preprint arXiv:2303.08774}.

\bibitem[{Adelani et~al.(2022)Adelani, Neubig, Ruder, Rijhwani, Beukman, Palen-Michel, Lignos, Alabi, Muhammad, Nabende et~al.}]{adelani2022masakhaner}
David Adelani, Graham Neubig, Sebastian Ruder, Shruti Rijhwani, Michael Beukman, Chester Palen-Michel, Constantine Lignos, Jesujoba Alabi, Shamsuddeen Muhammad, Peter Nabende, et~al. 2022.
\newblock Masakhaner 2.0: Africa-centric transfer learning for named entity recognition.
\newblock In \emph{Proceedings of the 2022 Conference on Empirical Methods in Natural Language Processing}, pages 4488--4508.

\bibitem[{Asai et~al.(2024)Asai, Kudugunta, Yu, Blevins, Gonen, Reid, Tsvetkov, Ruder, and Hajishirzi}]{asai-etal-2024-buffet}
Akari Asai, Sneha Kudugunta, Xinyan Yu, Terra Blevins, Hila Gonen, Machel Reid, Yulia Tsvetkov, Sebastian Ruder, and Hannaneh Hajishirzi. 2024.
\newblock \href {https://doi.org/10.18653/v1/2024.naacl-long.100} {{BUFFET}: Benchmarking large language models for few-shot cross-lingual transfer}.
\newblock In \emph{Proceedings of the 2024 Conference of the North American Chapter of the Association for Computational Linguistics: Human Language Technologies (Volume 1: Long Papers)}, pages 1771--1800, Mexico City, Mexico. Association for Computational Linguistics.

\bibitem[{Bogdanov et~al.(2024)Bogdanov, Constantin, Bernard, Crabb{\'e}, and Bernard}]{bogdanov2024nuner}
Sergei Bogdanov, Alexandre Constantin, Timoth{\'e}e Bernard, Benoit Crabb{\'e}, and Etienne Bernard. 2024.
\newblock Nuner: Entity recognition encoder pre-training via llm-annotated data.
\newblock \emph{arXiv e-prints}, pages arXiv--2402.

\bibitem[{Conneau et~al.(2020)Conneau, Khandelwal, Goyal, Chaudhary, Wenzek, Guzm{\'a}n, Grave, Ott, Zettlemoyer, and Stoyanov}]{conneau-etal-2020-unsupervised}
Alexis Conneau, Kartikay Khandelwal, Naman Goyal, Vishrav Chaudhary, Guillaume Wenzek, Francisco Guzm{\'a}n, Edouard Grave, Myle Ott, Luke Zettlemoyer, and Veselin Stoyanov. 2020.
\newblock \href {https://doi.org/10.18653/v1/2020.acl-main.747} {Unsupervised cross-lingual representation learning at scale}.
\newblock In \emph{Proceedings of the 58th Annual Meeting of the Association for Computational Linguistics}, pages 8440--8451, Online. Association for Computational Linguistics.

\bibitem[{Dang et~al.(2024)Dang, Singh, D'souza, Ahmadian, Salamanca, Smith, Peppin, Hong, Govindassamy, Zhao et~al.}]{dang2024aya}
John Dang, Shivalika Singh, Daniel D'souza, Arash Ahmadian, Alejandro Salamanca, Madeline Smith, Aidan Peppin, Sungjin Hong, Manoj Govindassamy, Terrence Zhao, et~al. 2024.
\newblock Aya expanse: Combining research breakthroughs for a new multilingual frontier.
\newblock \emph{arXiv preprint arXiv:2412.04261}.

\bibitem[{Dao et~al.(2025)Dao, Teranishi, Matsumoto, Boudin, and Aizawa}]{dao-etal-2025-overcoming}
An~Dao, Hiroki Teranishi, Yuji Matsumoto, Florian Boudin, and Akiko Aizawa. 2025.
\newblock \href {https://doi.org/10.18653/v1/2025.bionlp-1.28} {Overcoming data scarcity in named entity recognition: Synthetic data generation with large language models}.
\newblock In \emph{Proceedings of the 24th Workshop on Biomedical Language Processing}, pages 328--340, Viena, Austria. Association for Computational Linguistics.

\bibitem[{Dubey et~al.(2024)Dubey, Jauhri, Pandey, Kadian, Al-Dahle, Letman, Mathur, Schelten, Yang, Fan et~al.}]{dubey2024llama}
Abhimanyu Dubey, Abhinav Jauhri, Abhinav Pandey, Abhishek Kadian, Ahmad Al-Dahle, Aiesha Letman, Akhil Mathur, Alan Schelten, Amy Yang, Angela Fan, et~al. 2024.
\newblock The llama 3 herd of models.
\newblock \emph{arXiv preprint arXiv:2407.21783}.

\bibitem[{Evuru et~al.(2024)Evuru, Ghosh, Kumar, S, Tyagi, and Manocha}]{evuru-etal-2024-coda}
Chandra~Kiran Evuru, Sreyan Ghosh, Sonal Kumar, Ramaneswaran S, Utkarsh Tyagi, and Dinesh Manocha. 2024.
\newblock \href {https://doi.org/10.18653/v1/2024.findings-naacl.238} {{C}o{D}a: Constrained generation based data augmentation for low-resource {NLP}}.
\newblock In \emph{Findings of the Association for Computational Linguistics: NAACL 2024}, pages 3754--3769, Mexico City, Mexico. Association for Computational Linguistics.

\bibitem[{Fetahu et~al.(2022)Fetahu, Fang, Rokhlenko, and Malmasi}]{fetahu-etal-2022-dynamic}
Besnik Fetahu, Anjie Fang, Oleg Rokhlenko, and Shervin Malmasi. 2022.
\newblock \href {https://doi.org/10.18653/v1/2022.naacl-main.200} {Dynamic gazetteer integration in multilingual models for cross-lingual and cross-domain named entity recognition}.
\newblock In \emph{Proceedings of the 2022 Conference of the North American Chapter of the Association for Computational Linguistics: Human Language Technologies}, pages 2777--2790, Seattle, United States. Association for Computational Linguistics.

\bibitem[{Heng et~al.(2024)Heng, Deng, Li, Yu, Li, Zhang, and Zhang}]{heng-etal-2024-proggen}
Yuzhao Heng, Chunyuan Deng, Yitong Li, Yue Yu, Yinghao Li, Rongzhi Zhang, and Chao Zhang. 2024.
\newblock \href {https://doi.org/10.18653/v1/2024.findings-acl.947} {{P}rog{G}en: Generating named entity recognition datasets step-by-step with self-reflexive large language models}.
\newblock In \emph{Findings of the Association for Computational Linguistics: ACL 2024}, pages 15992--16030, Bangkok, Thailand. Association for Computational Linguistics.

\bibitem[{Krishnan et~al.(2021)Krishnan, Ziehe, Pannach, and Sporleder}]{krishnan2021employing}
Aravind Krishnan, Stefan Ziehe, Franziska Pannach, and Caroline Sporleder. 2021.
\newblock Employing wikipedia as a resource for named entity recognition in morphologically complex under-resourced languages.
\newblock In \emph{Proceedings of the 14th Workshop on Building and Using Comparable Corpora (BUCC 2021)}, pages 28--39.

\bibitem[{Kwon et~al.(2023)Kwon, Li, Zhuang, Sheng, Zheng, Yu, Gonzalez, Zhang, and Stoica}]{kwon2023efficient}
Woosuk Kwon, Zhuohan Li, Siyuan Zhuang, Ying Sheng, Lianmin Zheng, Cody~Hao Yu, Joseph Gonzalez, Hao Zhang, and Ion Stoica. 2023.
\newblock Efficient memory management for large language model serving with pagedattention.
\newblock In \emph{Proceedings of the 29th symposium on operating systems principles}, pages 611--626.

\bibitem[{Li et~al.(2023)Li, Zhu, Lu, and Yin}]{li2023synthetic}
Zhuoyan Li, Hangxiao Zhu, Zhuoran Lu, and Ming Yin. 2023.
\newblock Synthetic data generation with large language models for text classification: Potential and limitations.
\newblock In \emph{Proceedings of the 2023 Conference on Empirical Methods in Natural Language Processing}, pages 10443--10461.

\bibitem[{Lignos et~al.(2022)Lignos, Holley, Palen-Michel, and S{\"a}lev{\"a}}]{lignos-etal-2022-toward}
Constantine Lignos, Nolan Holley, Chester Palen-Michel, and Jonne S{\"a}lev{\"a}. 2022.
\newblock \href {https://doi.org/10.18653/v1/2022.findings-acl.44} {Toward more meaningful resources for lower-resourced languages}.
\newblock In \emph{Findings of the Association for Computational Linguistics: ACL 2022}, pages 523--532, Dublin, Ireland. Association for Computational Linguistics.

\bibitem[{Liu et~al.(2021)Liu, Ding, Bing, Joty, Si, and Miao}]{liu-etal-2021-mulda}
Linlin Liu, Bosheng Ding, Lidong Bing, Shafiq Joty, Luo Si, and Chunyan Miao. 2021.
\newblock \href {https://doi.org/10.18653/v1/2021.acl-long.453} {{M}ul{DA}: A multilingual data augmentation framework for low-resource cross-lingual {NER}}.
\newblock In \emph{Proceedings of the 59th Annual Meeting of the Association for Computational Linguistics and the 11th International Joint Conference on Natural Language Processing (Volume 1: Long Papers)}, pages 5834--5846, Online. Association for Computational Linguistics.

\bibitem[{Malmasi et~al.(2022)Malmasi, Fang, Fetahu, Kar, and Rokhlenko}]{malmasi2022multiconer}
Shervin Malmasi, Anjie Fang, Besnik Fetahu, Sudipta Kar, and Oleg Rokhlenko. 2022.
\newblock Multiconer: A large-scale multilingual dataset for complex named entity recognition.
\newblock In \emph{Proceedings of the 29th International Conference on Computational Linguistics}, pages 3798--3809.

\bibitem[{Mayhew et~al.(2024)Mayhew, Blevins, Liu, {\v{S}}uppa, Gonen, Imperial, Karlsson, Lin, Ljube{\v{s}}i{\'c}, Miranda et~al.}]{mayhew2024universal}
Stephen Mayhew, Terra Blevins, Shuheng Liu, Marek {\v{S}}uppa, Hila Gonen, Joseph~Marvin Imperial, B{\"o}rje Karlsson, Peiqin Lin, Nikola Ljube{\v{s}}i{\'c}, Lester~James Miranda, et~al. 2024.
\newblock Universal ner: A gold-standard multilingual named entity recognition benchmark.
\newblock In \emph{Proceedings of the 2024 Conference of the North American Chapter of the Association for Computational Linguistics: Human Language Technologies (Volume 1: Long Papers)}, pages 4322--4337.

\bibitem[{Mayhew et~al.(2017)Mayhew, Tsai, and Roth}]{mayhew-etal-2017-cheap}
Stephen Mayhew, Chen-Tse Tsai, and Dan Roth. 2017.
\newblock \href {https://doi.org/10.18653/v1/D17-1269} {Cheap translation for cross-lingual named entity recognition}.
\newblock In \emph{Proceedings of the 2017 Conference on Empirical Methods in Natural Language Processing}, pages 2536--2545, Copenhagen, Denmark. Association for Computational Linguistics.

\bibitem[{Mhaske et~al.(2023)Mhaske, Kedia, Doddapaneni, Khapra, Kumar, Murthy, and Kunchukuttan}]{mhaske-etal-2023-naamapadam}
Arnav Mhaske, Harshit Kedia, Sumanth Doddapaneni, Mitesh~M. Khapra, Pratyush Kumar, Rudra Murthy, and Anoop Kunchukuttan. 2023.
\newblock \href {https://doi.org/10.18653/v1/2023.acl-long.582} {Naamapadam: A large-scale named entity annotated data for {I}ndic languages}.
\newblock In \emph{Proceedings of the 61st Annual Meeting of the Association for Computational Linguistics (Volume 1: Long Papers)}, pages 10441--10456, Toronto, Canada. Association for Computational Linguistics.

\bibitem[{Mueller et~al.(2020)Mueller, Andrews, and Dredze}]{mueller-etal-2020-sources}
David Mueller, Nicholas Andrews, and Mark Dredze. 2020.
\newblock \href {https://doi.org/10.18653/v1/2020.acl-main.720} {Sources of transfer in multilingual named entity recognition}.
\newblock In \emph{Proceedings of the 58th Annual Meeting of the Association for Computational Linguistics}, pages 8093--8104, Online. Association for Computational Linguistics.

\bibitem[{Ni et~al.(2017)Ni, Dinu, and Florian}]{ni-etal-2017-weakly}
Jian Ni, Georgiana Dinu, and Radu Florian. 2017.
\newblock \href {https://doi.org/10.18653/v1/P17-1135} {Weakly supervised cross-lingual named entity recognition via effective annotation and representation projection}.
\newblock In \emph{Proceedings of the 55th Annual Meeting of the Association for Computational Linguistics (Volume 1: Long Papers)}, pages 1470--1480, Vancouver, Canada. Association for Computational Linguistics.

\bibitem[{OpenAI(2025)}]{openai2025gpt4_1}
OpenAI. 2025.
\newblock Introducing {GPT-4.1} in the api.
\newblock \url{https://openai.com/index/gpt-4-1/}.
\newblock Accessed: 2025-11.

\bibitem[{Pan et~al.(2017)Pan, Zhang, May, Nothman, Knight, and Ji}]{pan-etal-2017-cross}
Xiaoman Pan, Boliang Zhang, Jonathan May, Joel Nothman, Kevin Knight, and Heng Ji. 2017.
\newblock \href {https://doi.org/10.18653/v1/P17-1178} {Cross-lingual name tagging and linking for 282 languages}.
\newblock In \emph{Proceedings of the 55th Annual Meeting of the Association for Computational Linguistics (Volume 1: Long Papers)}, pages 1946--1958, Vancouver, Canada. Association for Computational Linguistics.

\bibitem[{Rahimi et~al.(2019)Rahimi, Li, and Cohn}]{rahimi-etal-2019-massively}
Afshin Rahimi, Yuan Li, and Trevor Cohn. 2019.
\newblock \href {https://doi.org/10.18653/v1/P19-1015} {Massively multilingual transfer for {NER}}.
\newblock In \emph{Proceedings of the 57th Annual Meeting of the Association for Computational Linguistics}, pages 151--164, Florence, Italy. Association for Computational Linguistics.

\bibitem[{Schmidt et~al.(2022)Schmidt, Vuli{\'c}, and Glava{\v{s}}}]{schmidt-etal-2022-slicer}
Fabian~David Schmidt, Ivan Vuli{\'c}, and Goran Glava{\v{s}}. 2022.
\newblock \href {https://doi.org/10.18653/v1/2022.emnlp-main.740} {{SLICER}: Sliced fine-tuning for low-resource cross-lingual transfer for named entity recognition}.
\newblock In \emph{Proceedings of the 2022 Conference on Empirical Methods in Natural Language Processing}, pages 10775--10785, Abu Dhabi, United Arab Emirates. Association for Computational Linguistics.

\bibitem[{Whitehouse et~al.(2023)Whitehouse, Choudhury, and Aji}]{whitehouse-etal-2023-llm}
Chenxi Whitehouse, Monojit Choudhury, and Alham~Fikri Aji. 2023.
\newblock \href {https://doi.org/10.18653/v1/2023.emnlp-main.44} {{LLM}-powered data augmentation for enhanced cross-lingual performance}.
\newblock In \emph{Proceedings of the 2023 Conference on Empirical Methods in Natural Language Processing}, pages 671--686, Singapore. Association for Computational Linguistics.

\bibitem[{Xie et~al.(2018)Xie, Yang, Neubig, Smith, and Carbonell}]{xie-etal-2018-neural}
Jiateng Xie, Zhilin Yang, Graham Neubig, Noah~A. Smith, and Jaime Carbonell. 2018.
\newblock \href {https://doi.org/10.18653/v1/D18-1034} {Neural cross-lingual named entity recognition with minimal resources}.
\newblock In \emph{Proceedings of the 2018 Conference on Empirical Methods in Natural Language Processing}, pages 369--379, Brussels, Belgium. Association for Computational Linguistics.

\bibitem[{Yang et~al.(2022)Yang, Huang, Ma, Yin, Dong, Zhang, Guo, Li, and Wei}]{yang-etal-2022-crop}
Jian Yang, Shaohan Huang, Shuming Ma, Yuwei Yin, Li~Dong, Dongdong Zhang, Hongcheng Guo, Zhoujun Li, and Furu Wei. 2022.
\newblock \href {https://doi.org/10.18653/v1/2022.findings-emnlp.34} {{CROP}: Zero-shot cross-lingual named entity recognition with multilingual labeled sequence translation}.
\newblock In \emph{Findings of the Association for Computational Linguistics: EMNLP 2022}, pages 486--496, Abu Dhabi, United Arab Emirates. Association for Computational Linguistics.

\bibitem[{Zeng et~al.(2022)Zeng, Jiang, Yin, Wang, Lin, and Cao}]{zeng-etal-2022-dualner}
Jiali Zeng, Yufan Jiang, Yongjing Yin, Xu~Wang, Binghuai Lin, and Yunbo Cao. 2022.
\newblock \href {https://doi.org/10.18653/v1/2022.findings-emnlp.132} {{D}ual{NER}: A dual-teaching framework for zero-shot cross-lingual named entity recognition}.
\newblock In \emph{Findings of the Association for Computational Linguistics: EMNLP 2022}, pages 1837--1843, Abu Dhabi, United Arab Emirates. Association for Computational Linguistics.

\bibitem[{Zhao et~al.(2022)Zhao, Du, Liu, and Zhu}]{zhao-etal-2022-transadv}
Yichun Zhao, Jintao Du, Gongshen Liu, and Huijia Zhu. 2022.
\newblock \href {https://doi.org/10.18653/v1/2022.findings-emnlp.52} {{T}rans{A}dv: A translation-based adversarial learning framework for zero-resource cross-lingual named entity recognition}.
\newblock In \emph{Findings of the Association for Computational Linguistics: EMNLP 2022}, pages 742--749, Abu Dhabi, United Arab Emirates. Association for Computational Linguistics.

\bibitem[{Zhou et~al.(2022)Zhou, Li, Bing, Cambria, Si, and Miao}]{zhou-etal-2022-conner}
Ran Zhou, Xin Li, Lidong Bing, Erik Cambria, Luo Si, and Chunyan Miao. 2022.
\newblock \href {https://doi.org/10.18653/v1/2022.emnlp-main.577} {{C}on{NER}: Consistency training for cross-lingual named entity recognition}.
\newblock In \emph{Proceedings of the 2022 Conference on Empirical Methods in Natural Language Processing}, pages 8438--8449, Abu Dhabi, United Arab Emirates. Association for Computational Linguistics.

\end{thebibliography}

\appendix

\section{Synthetic Data Generation}
\label{app:datagen}

As shown in Figure \ref{fig:datagen}, we present the following prompt to the LLM in the data generation process:
 \\ 
 
\texttt{Help me make a \{language\} Named Entity Recognition dataset. Please give me \{n\} new datapoints, formatted as a single JSON object. Make sure the examples are unique and diverse. Here are some examples to get you started:}

\texttt{\{m examples\}} \\

We prompt \texttt{GPT-4.1} using OpenAI's batch API functionality\footnote{https://platform.openai.com/docs/guides/batch}; for the open-source models, we use the \texttt{vLLM} library \cite{kwon2023efficient} to run inference.

For \texttt{GPT-4.1}, we used the OpenAI API's functionalities for structured outputs to ensure that outputs were formatted as JSON strings. For the open-sourced models, we experimented with using \texttt{transformers}-compatible libraries for obtaining structured outputs from LLMs, but ultimately found better results simply specifying the JSON requirement in the model and system prompt.  For the open-sourced models, we used the following system prompt:\\

\texttt{You are a helpful model that helps build text-based datasets, but does not produce any conversation besides the text it is asked to produce. You only output JSON strings.}\\

For \texttt{GPT-4.1}, we used the following (minimally different) system prompt, on the assumption that specifying output mode in the system prompt was less important on account of the API's structured output functionalities:\\

\texttt{You are a helpful model that helps build text-based datasets, but does not produce any conversation besides the text it is asked to produce.}\\

We ran both open-sourced models with a temperature setting of 0.8, and nucleus sampling value of 0.8.
We initially used a maximum new token limit of 4096 for both models. However, noticing that some of \texttt{Llama-3.1-8B-Instruct}'s unusable datapoints were specifically due to hitting new token limits, we regenerated data from this model with a maximum new token limit of 8192.
Calls to \texttt{GPT-4.1} were made using default hyperparameters. 

Table~\ref{tab:responsequality} shows some of the examples of the different types of responses to these prompts.

\section{Related-Language Model Details}
\label{app:appendix_relatedlanguagemodel}

In the \textsc{ner fine-tuning} setting, we first train an NER model on a language related to the target language, before fine-tuning it further on the target language NER data.
Below is the list of related languages chosen to build a base NER model for each target language.

\begin{table}[htb!]
    \centering
    \begin{tabular}{@{}lr@{}}
    \toprule
       Target Language  &  Related Language Chosen \\ \midrule
        Kannada & Telugu \\
        Tamil & Telugu \\
        Telugu & Kannada \\
        Malayalam & Tamil \\ 
        Kinyarwanda & Swahili \\
        Swahili & Kinyarwanda \\
        Yoruba & Igbo \\
        Igbo & Yoruba \\
        Swedish & Danish \\
        Danish & Swedish \\
        Slovak & English* \\
        \bottomrule
    \end{tabular}
    \caption{List of related languages used in the \textsc{ner fine-tuning} setting for each target language. *English is not closely related to Slovak, but given the absence of another closely related language among the 11 target languages, it was chosen as the language for the base NER model to be fine-tuned.}
    \label{tab:relatedlangappendix}
\end{table}

\subsection{NER-fine tuning: Implementation Details}
\label{app:finetuningimplementation}
We source the pre-trained \texttt{XLM-RoBERTa-large} weights from Huggingface using the \texttt{transformers} library; fine-tuning is implemented using training pipelines from the same library.  In the \textsc{ner from scratch} setting, we train on the the target language data for 10 epochs; in the \textsc{ner fine-tuning} setting, we train on the related language data for 5 epochs, and then the target language data for 10 epochs. In all cases, we use a learning rate of 2e-05, and a batch size of 16.

\section{Full Results of WikiANN Comparison}
\label{app:wikianndetails}
The WikiANN dataset is a massively multilingual NER benchmark, comprising data from 176 languages \cite{pan-etal-2017-cross, rahimi-etal-2019-massively}.\footnote{As \citet{lignos-etal-2022-toward} also note, strictly speaking, the original version of WikiANN put together by \citet{pan-etal-2017-cross} contains data from 282 languages; the version of the dataset commonly downloaded from Huggingface, however, and put together by \citet{rahimi-etal-2019-massively}, contains data from 176 languages. In this work, we refer to the latter when referring to the WikiANN dataset.} 
Table \ref{tab:appendix_wikiann} shows the full list of comparisons between NER model performance when trained on organic data, \texttt{GPT-4.1}-produced data, and WikiANN data. 
The sizes of the WikiANN train sets vary significantly between different languages, meaning we often cannot assess the quality of the data in the context of training sets containing over 1000 datapoints (e.g. Kannada and Yoruba, whose WikiANN train sets contain only 100 datapoints). 
In such cases, however, we compare model performance when trained on equally small amounts of organic or LLM-produced synthetic data.

\begin{table}[t]
\resizebox{\columnwidth}{!}{
    \begin{tabular}{@{}llccc@{}}
    \toprule
       Language &  &  \textsc{n.f.s.} F1 & \textsc{n.f.t.} F1 & \textsc{Data Size} \\ \midrule
        Kannada & \textsc{WikiANN} & 4.5e-3 & 0.77 & 100 \\ 
        & \textsc{GPT-4.1} & 0.03 & 0.75 & 100 \\ 
        & \textsc{GPT-4.1} & 0.00 & 0.43 & 4899 \\ 
        & \textsc{Naamapadam} & 0.49 & 0.79 & 100 \\ 
        & \textsc{Naamapadam} & \textbf{0.80} & \textbf{0.78} & 5000 \\ \midrule
        Telugu & \textsc{WikiANN} & 0.67 & 0.74 & 1000 \\ 
        & \textsc{GPT-4.1} & 0.00 & 0.40 & 1000 \\ 
        & \textsc{GPT-4.1} & 0.18 & 0.47 & 4931 \\ 
        & \textsc{Naamapadam} & 0.71 & 0.81 & 1000 \\ 
        & \textsc{Naamapadam} & \textbf{0.84} & \textbf{0.83} & 5000 \\ \midrule
        Tamil & \textsc{WikiANN} & 0.55 & 0.62 & 15000 \\ 
        & \textsc{GPT-4.1} & 0.00 & 0.57 & 4944 \\ 
        & \textsc{Naamapadam} & \textbf{0.73} & \textbf{0.74} & 5000 \\ \midrule
        Malayalam & \textsc{WikiANN} & 0.65 & 0.74 & 10000 \\ 
        & \textsc{GPT-4.1} & 0.25 & 0.35 & 4859 \\ 
        & \textsc{Naamapadam} & \textbf{0.83} & \textbf{0.83} & 5000 \\ \midrule
        Yoruba & \textsc{WikiANN} & 0.07 & 0.21 & 100 \\ 
        & \textsc{GPT-4.1} & 0.20 & 0.42 & 100 \\
        & \textsc{GPT-4.1} & 0.37 & 0.50 & 4187 \\
        & \textsc{MasakhaNER 2} & 0.28 & 0.54 & 100 \\ 
        & \textsc{MasakhaNER 2} & \textbf{0.83} & \textbf{0.84} & 5000 \\ \midrule
        Swahili & \textsc{WikiANN} & 0.50 & 0.59 & 1000 \\ 
        & \textsc{GPT-4.1} & 0.74 & 0.78 & 1000 \\ 
        & \textsc{GPT-4.1} & 0.78 & 0.79 & 2619 \\ 
        & \textsc{MasakhaNER 2} & 0.88 & 0.91 & 1000 \\
        & \textsc{MasakhaNER 2} & \textbf{0.92} & \textbf{0.91} & 5000 \\ \midrule
        Kinyarwanda & \textsc{WikiANN} & 7.9e-4 & 0.35 & 100 \\ 
        & \textsc{GPT-4.1} & 0.29 & 0.46 & 100 \\ 
        & \textsc{GPT-4.1} & 0.47 & 0.50 & 2683 \\ 
        & \textsc{MasakhaNER 2} & 0.07 & 0.61 & 100 \\
        & \textsc{MasakhaNER 2} & \textbf{0.79} & \textbf{0.81} & 5000 \\ \midrule
        Igbo & \textsc{WikiANN} & 7.7e-3 & 0.39 & 100 \\ 
        & \textsc{GPT-4.1} & 0.29 & 0.55 & 100 \\ 
        & \textsc{GPT-4.1} & 0.53 & 0.55 & 3473 \\ 
        & \textsc{MasakhaNER 2} & 0.48 & 0.76 & 100 \\
        & \textsc{MasakhaNER 2} & \textbf{0.82} & \textbf{0.83} & 5000 \\ \midrule
        Danish & \textsc{WikiANN} & 0.72 & 0.71 & 20000 \\ 
        & \textsc{GPT-4.1} & 0.79 & 0.82 & 4112 \\ 
        & \textsc{Universal NER} & \textbf{0.83} & \textbf{0.85} & 4383 \\ \midrule
        Swedish & \textsc{WikiANN} & 0.36 & 0.29 & 20000 \\ 
        & \textsc{GPT-4.1} & 0.69 & 0.74 & 4215 \\ 
        & \textsc{Universal NER} & \textbf{0.87} & \textbf{0.90} & 4303 \\ \midrule
        Slovak & \textsc{WikiANN} & 0.57 & 0.55 & 20000 \\ 
        & \textsc{GPT-4.1} & 0.29 & 0.29 & 4508 \\ 
        & \textsc{Universal NER} & \textbf{0.77} & \textbf{0.80} & 5000 \\ \midrule
        \bottomrule
    \end{tabular}
    }
    \caption{Performance of NER models trained on WikiANN, synthetic data from \texttt{GPT-4.1}, and high quality `organic' data, for all 11 languages. \textsc{n.f.s}: \textsc{ner from scratch} setting; \textsc{n.f.t}: \textsc{ner fine-tuning} setting.}
    \label{tab:appendix_wikiann}
\end{table}

\end{document}